\documentclass{article}

\usepackage{microtype}
\usepackage{graphicx}
\usepackage{subfigure}
\usepackage{booktabs} 

\usepackage{hyperref}

\usepackage[accepted]{icml2024}

\usepackage{amsmath}
\usepackage{amssymb}
\usepackage{mathtools}
\usepackage{amsthm}

\usepackage[capitalize,noabbrev]{cleveref}

\usepackage{inconsolata}
\usepackage{placeins}
\usepackage{listings}

\theoremstyle{plain}

\theoremstyle{definition}

\theoremstyle{remark}

\usepackage[textsize=tiny]{todonotes}

\usepackage{listings}

\usepackage[utf8]{inputenc}
\usepackage[T1]{fontenc}
\usepackage{hyphenat}
\usepackage{xspace}
\usepackage{amsmath}
\usepackage{amsfonts}
\usepackage{hyperref}
\usepackage{url}
\usepackage{booktabs}
\usepackage{multirow}
\usepackage{makecell}
\usepackage{minibox}
\usepackage{bbm}
\usepackage{graphicx}
\usepackage{balance}
\usepackage{mathtools}
\usepackage{color}
\usepackage{marvosym}
\usepackage{ifthen}
\usepackage{textcomp}
\usepackage{enumitem}
\usepackage{verbatim}
\usepackage{algorithm}
\usepackage{numprint}
\usepackage{balance}

\usepackage{amsthm}
\theoremstyle{plain}

\newcommand{\chatoDisplayMode}[1]{#1}

\definecolor{MyRed}{rgb}{0.6,0.0,0.0} 
\definecolor{MyBlack}{rgb}{0.1,0.1,0.1} 
\newcommand{\inred}[1]{{\color{MyRed}\sf\textbf{\textsc{#1}}}}
\newcommand{\frameit}[2]{
  \begin{center}
  {\color{MyRed}
  \framebox[.9\columnwidth][l]{
    \begin{minipage}{.85\columnwidth}
    \inred{#1}: {\sf\color{MyBlack}#2}
    \end{minipage}
  }\\
  }
  \end{center}
}

\newcommand{\note}[2][]{\chatoDisplayMode{\def\@tmpsig{#1}\frameit{{\Pointinghand} Note}{#2\ifx \@tmpsig \@empty \else \mbox{ --\em #1}\fi}}}

\newcommand{\abbrevStyle}[1]{#1}

\usepackage{amssymb}%
\usepackage{pifont}%

\newcommand{\xmark}{\ding{55}}
\newcommand{\ie}{\abbrevStyle{i.e.}\xspace}
\newcommand{\eg}{\abbrevStyle{e.g.}\xspace}
\newcommand{\cf}{\abbrevStyle{cf.}\xspace}

\newcommand{\etc}{\abbrevStyle{etc.}\xspace}

\newcommand{\Secref}[1]{Sec.~\ref{#1}}

\newcommand{\Tabref}[1]{Table~\ref{#1}}
\newcommand{\Figref}[1]{Fig.~\ref{#1}}

\newcommand{\Appref}[1]{Appendix~\ref{#1}}

\newcommand{\xhdr}[1]{\vspace{1.7mm}\noindent{{\bf #1.}}}

\newcommand{\xhdrNoPeriod}[1]{\vspace{1.7mm}\noindent{{\bf #1}}}

\newcommand{\denselist}{ \itemsep -2pt\topsep-10pt\partopsep-10pt }

\newcommand{\textcite}[1]{\citeauthor{#1} \shortcite{#1}}

\newcommand{\hide}[1]{}

\makeatletter
\newcommand{\iffont}[2]{\ifthenelse{\equal{\f@family}{#1}}{#2}{}}
\makeatother

  \usepackage{mathptmx}

  \DeclareSymbolFont{greek}{OML}{cmm}{m}{n}
  \DeclareMathSymbol{\alpha}{\mathalpha}{greek}{"0B}
  \DeclareMathSymbol{\beta}{\mathalpha}{greek}{"0C}
  \DeclareMathSymbol{\gamma}{\mathalpha}{greek}{"0D}
  \DeclareMathSymbol{\delta}{\mathalpha}{greek}{"0E}
  \DeclareMathSymbol{\epsilon}{\mathalpha}{greek}{"0F}
  \DeclareMathSymbol{\zeta}{\mathalpha}{greek}{"10}
  \DeclareMathSymbol{\eta}{\mathalpha}{greek}{"11}
  \DeclareMathSymbol{\theta}{\mathalpha}{greek}{"12}
  \DeclareMathSymbol{\iota}{\mathalpha}{greek}{"13}
  \DeclareMathSymbol{\kappa}{\mathalpha}{greek}{"14}
  \DeclareMathSymbol{\lambda}{\mathalpha}{greek}{"15}
  \DeclareMathSymbol{\mu}{\mathalpha}{greek}{"16}
  \DeclareMathSymbol{\nu}{\mathalpha}{greek}{"17}
  \DeclareMathSymbol{\xi}{\mathalpha}{greek}{"18}
  \DeclareMathSymbol{\pi}{\mathalpha}{greek}{"19}
  \DeclareMathSymbol{\rho}{\mathalpha}{greek}{"1A}
  \DeclareMathSymbol{\sigma}{\mathalpha}{greek}{"1B}
  \DeclareMathSymbol{\tau}{\mathalpha}{greek}{"1C}
  \DeclareMathSymbol{\upsilon}{\mathalpha}{greek}{"1D}
  \DeclareMathSymbol{\phi}{\mathalpha}{greek}{"1E}
  \DeclareMathSymbol{\chi}{\mathalpha}{greek}{"1F}
  \DeclareMathSymbol{\psi}{\mathalpha}{greek}{"20}
  \DeclareMathSymbol{\omega}{\mathalpha}{greek}{"21}
  \DeclareMathSymbol{\varepsilon}{\mathalpha}{greek}{"22}
  \DeclareMathSymbol{\vartheta}{\mathalpha}{greek}{"23}
  \DeclareMathSymbol{\varpi}{\mathalpha}{greek}{"24}
  \DeclareMathSymbol{\varrho}{\mathalpha}{greek}{"25}
  \DeclareMathSymbol{\varsigma}{\mathalpha}{greek}{"26}
  \DeclareMathSymbol{\varphi}{\mathalpha}{greek}{"27}
  \DeclareSymbolFont{otone}{OT1}{cmr}{m}{n}
  \DeclareMathSymbol{\Gamma}{\mathalpha}{otone}{0}
  \DeclareMathSymbol{\Delta}{\mathalpha}{otone}{1}
  \DeclareMathSymbol{\Theta}{\mathalpha}{otone}{2}
  \DeclareMathSymbol{\Lambda}{\mathalpha}{otone}{3}
  \DeclareMathSymbol{\Xi}{\mathalpha}{otone}{4}
  \DeclareMathSymbol{\Pi}{\mathalpha}{otone}{5}
  \DeclareMathSymbol{\Sigma}{\mathalpha}{otone}{6}
  \DeclareMathSymbol{\Upsilon}{\mathalpha}{otone}{7}
  \DeclareMathSymbol{\Phi}{\mathalpha}{otone}{8}
  \DeclareMathSymbol{\Psi}{\mathalpha}{otone}{9}
  \DeclareMathSymbol{\Omega}{\mathalpha}{otone}{10}
  \DeclareSymbolFont{syms}{OML}{cmm}{m}{it}
  \DeclareMathSymbol{\partial}{\mathord}{syms}{"40}
  \DeclareMathAlphabet{\mathbold}{OML}{cmm}{b}{it}
  \DeclareSymbolFont{largesymbols}{OMX}{cmex}{m}{n}

  \DeclareMathAlphabet{\mathcal}{OMS}{cmsy}{m}{n}

\newcommand{\libraryName}{\texorpdfstring{$\mathtt{aiFlows}$}}
\newcommand{\frameworkName}{\textit{Flows}}

\newcommand{\gpt}{GPT-4}

\DeclareSymbolFont{extraup}{U}{zavm}{m}{n}
\DeclareMathSymbol{\microsoft}{\mathalpha}{extraup}{81}
\DeclareMathSymbol{\epfl}{\mathalpha}{extraup}{83}
\DeclareMathSymbol{\psl}{\mathalpha}{extraup}{84}

\icmltitlerunning{Submission and Formatting Instructions for ICML 2024}

\begin{document}

\twocolumn[
\icmltitle{Flows: Building Blocks of Reasoning and Collaborating AI}



\icmlsetsymbol{epfl}{1}
\icmlsetsymbol{equal}{*}
\icmlsetsymbol{equall}{**}
\icmlsetsymbol{grenoble}{2}
\icmlsetsymbol{psl}{3}

\begin{icmlauthorlist}
\icmlauthor{Martin Josifoski}{equal, epfl}
\icmlauthor{Lars Klein}{equal, epfl}
\icmlauthor{Maxime Peyrard}{grenoble}
\icmlauthor{Nicolas Baldwin}{epfl}
\icmlauthor{Yifei Li}{equall, epfl}
\icmlauthor{Saibo Geng}{equall, epfl}
\icmlauthor{Julian Paul Schnitzler}{epfl}
\icmlauthor{Yuxing Yao}{epfl}
\icmlauthor{Jiheng Wei}{psl}
\icmlauthor{Debjit Paul}{epfl}
\icmlauthor{Robert West}{epfl}
\end{icmlauthorlist}


\icmlcorrespondingauthor{Martin Josifoski}{martin.josifoski@epfl.ch}
\icmlcorrespondingauthor{Lars Klein}{lars.klein@epfl.ch} 
\icmlcorrespondingauthor{Maxime Peyrard}{maxime.peyrard@univ-grenoble-alpes.fr} 
\icmlcorrespondingauthor{Robert West}{robert.west@epfl.ch}

\icmlkeywords{Machine Learning, ICML}

\vskip 0.3in
]




\begin{abstract}
Recent advances in artificial intelligence (AI) have produced highly capable and controllable systems. This creates unprecedented opportunities for structured reasoning as well as collaboration among multiple AI systems and humans. 
To fully realize this potential, it is essential to develop a principled way of designing and studying such structured interactions.
For this purpose, we introduce the conceptual framework \frameworkName{}.
Flows are self-contained building blocks of computation, with an isolated state, communicating through a standardized message-based interface. 
This modular design simplifies the process of creating Flows by allowing them to be recursively composed into arbitrarily nested interactions and is inherently concurrency-friendly. 
Crucially, any interaction can be implemented using this framework, including prior work on AI--AI and human--AI interactions, prompt engineering schemes, and tool augmentation.
We demonstrate the potential of \frameworkName{} on competitive coding, a challenging task on which even \gpt{} struggles. 
Our results suggest that structured reasoning and collaboration substantially improve generalization, with AI-only Flows adding +$21$ and human--AI Flows adding +$54$ absolute points in terms of solve rate. 
To support rapid and rigorous research, we introduce the \libraryName{} library embodying \frameworkName{}. The \libraryName{} library is available at 
\url{https://github.com/epfl-dlab/aiflows}.
Data and Flows for reproducing our experiments are available at \url{https://github.com/epfl-dlab/cc_flows}.
\end{abstract}

\label{submission}

\section{Introduction}
\label{sec:introduction}

\begingroup
\renewcommand{\thefootnote}{} 
\footnotetext{\textsuperscript{1}EPFL \textsuperscript{2}Univ. Grenoble Alpes, CNRS, Grenoble INP, LIG \textsuperscript{3}PSL University *, ** Equal contribution \\Correspondence to: martin.josifoski@epfl.ch, lars.klein@epfl.ch, maxime.peyrard@univ-grenoble-alpes.fr, robert.west@epfl.ch}
\renewcommand{\thefootnote}{\arabic{footnote}} 
\endgroup

The success of large language models (LLMs) largely lies in their remarkable emergent ability to adapt to information within their context (\ie, prompt) \cite{NEURIPS2020_1457c0d6, wei2022chain, NEURIPS2022_8bb0d291}. By strategically crafting the context, LLMs can be conditioned to perform complex reasoning \cite{wei2022chain, DBLP:journals/corr/abs-2112-00114} and effectively utilize external tools \cite{Schick2023ToolformerLM}, significantly enhancing their capabilities. Some of the most exciting recent developments involve defining \textit{control flows}, wherein LLMs, with the ability to control a set of tools, are called in an orchestrated fashion to solve increasingly complex tasks. Examples of such control flows include ReAct \citep{yao2023react}, AutoGPT \citep{autogpt2023}, BabyAGI \citep{babyagi2023}, PromptBreeder \citep{DBLP:journals/corr/abs-2309-16797} and FunSearch \citep{romera2023mathematical}. Even the ubiquitous ChatGPT \citep{openai_chatgpt} application is an instance of a control flow built around the GPT-3.5 and GPT-4 models \cite{NEURIPS2020_1457c0d6, OpenAI2023GPT4TR}. However, these represent but a few of the many conceivable control flows, offering only a glimpse into the vast potential of structured LLM interactions. To realize this potential, we need to develop ways to study such interactions systematically.

\begin{figure*}[t]
    \centering   
    \includegraphics[scale=1, width=0.85\textwidth]{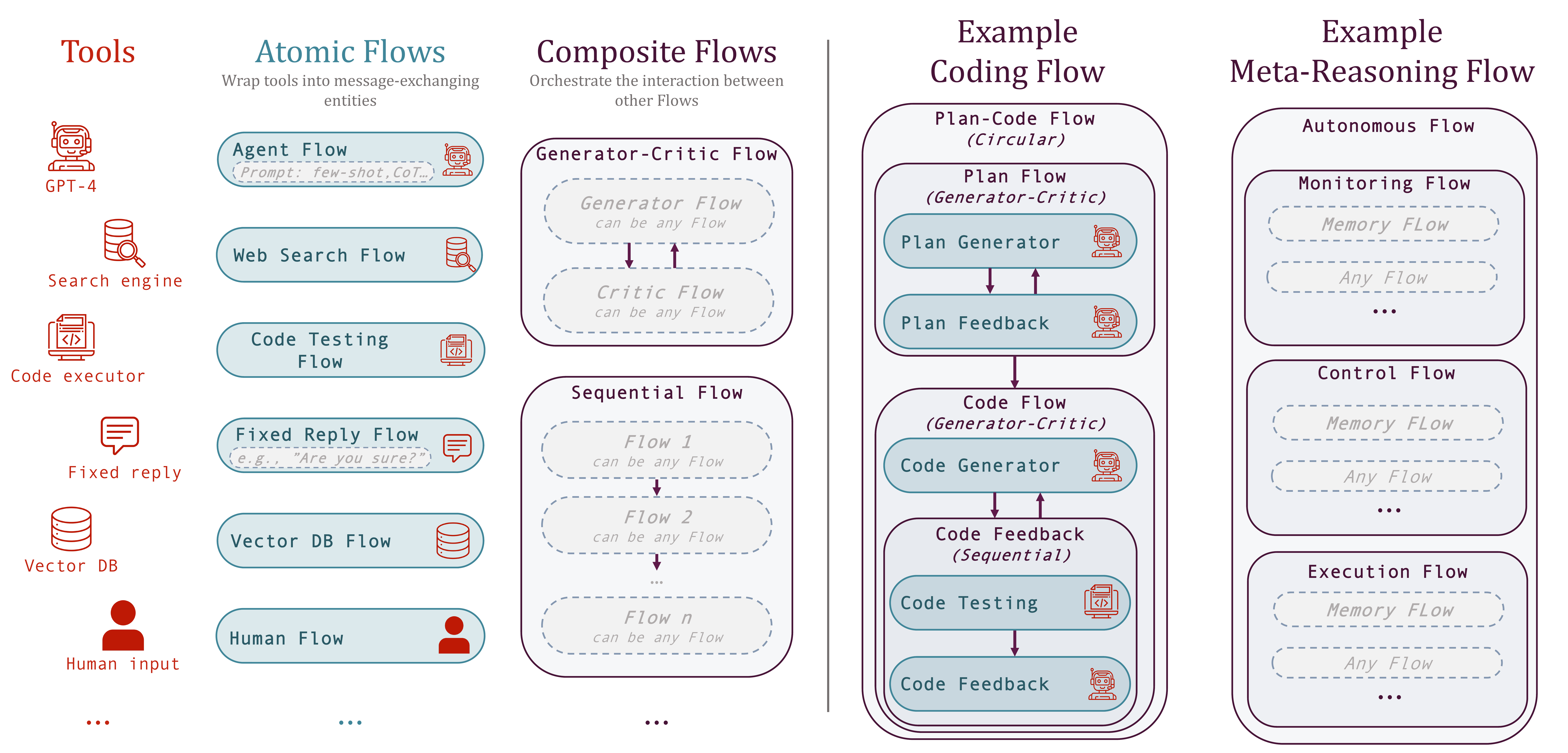}
    \caption[]{\textbf{\frameworkName{} framework exemplified.} The first column depicts examples of tools. The second column depicts Atomic Flows constructed from the example tools. The third column depicts examples of Composite Flows defining structured interaction between Atomic or Composite Flows. The fourth column illustrates a specific Composite competitive coding Flow as those used in the experiments. The fifth column outlines the structure of a hypothetical Flow, defining a meta-reasoning process that could support autonomous behavior.\footnotemark{}}
    \label{fig:flows}
\end{figure*}

In software engineering, simple processes can be implemented in an unstructured fashion, perhaps in a single file. However, as the size and complexity of the systems increase, choosing the right abstractions and architecture becomes critical \citep{DBLP:books/ws/93/GarlanS93}. Currently, for structured LLM interactions we want to model, implement, and study, we are at a point where this become unwieldy. Yet, no general efficient abstraction exists for effectively modeling arbitrarily complex structured interactions.
Previous work and existing frameworks, such as LangChain \cite{langchain2022}, Chameleon \cite{Lu2023ChameleonPC}, and HuggingGPT \cite{Shen2023HuggingGPTSA}, have converged to an ad-hoc abstraction that models agents as entities that use LLMs to select and execute actions towards specific tasks, where the set of possible actions is pre-defined by the available tools. In this view, tools serve a narrow, well-defined goal and can perform sophisticated tasks (\eg, querying a search engine or executing code). However, their behavior is {limited} to a single interaction. 
To highlight the implications of this limitation, consider the following scenario: Alice wants to apply for a job at HappyCorp. If Alice is an agent, she would need to explicitly plan the entire process, including preparing the application, sending it, and evaluating it, which may involve a background check, organizing interviews, and more. Alice would need the knowledge and the ``computational'' ability to account for every detail, including unforeseen events that may arise (\eg, the interviewer being on parental leave), and require her to adapt. 
In reality, most of the complexity is hidden from Alice behind an interface to HappyCorp's hiring process that might itself be composed of sub-processes involving many other \textit{agents} and \textit{tools}. 
Therefore, Alice is completely agnostic to the process(es) happening behind the interface and the respective logistics.
On the other hand, the hiring process, carefully designed by experts, can be reused by many agents, and its sub-processes can be modified or improved with minimal or no impact on the other components beyond an updated interface. 
This makes it evident that agents and tools should be able to interact in complex, dynamic or static, ways as parts of nested, modular processes (that run locally or remotely), and the distinction between the two becomes blurred as they both serve as computational units in a complex computational process. 

Starting from the observation that all processes are (control) flows defining a potentially complex interaction between many diverse components; we introduce a conceptual framework where Flows are the fundamental building blocks of computation. Flows are independent, self-contained, goal-driven entities able to complete semantically meaningful units of work. To exchange information, Flows communicate via a standardized message-based interface. The framework is depicted in \Figref{fig:flows}. 

The \frameworkName{} abstraction ensures modularity. Alice, a higher-level meta-reasoning Flow that can support autonomous behavior, does not need to know anything beyond how to interface with HappyCorp's hiring Flow. This substantially reduces complexity (Alice is interacting with a deeply nested, compositional structured interaction through a simple interface) and provides flexibility, allowing sub-Flows to be swapped without consequences as long as they have the same interface. Indeed, HappyCorp's pre-filtering Flow can be swapped from a rule-based system to an AI model or even a human Flow without affecting the structure of the overall process. The abstraction also enables reusability and the composition of sub-Flows into new Flows for different tasks. Furthermore, the framework shares key design choices with the Actor model, one of the most prominent models of concurrent computation (\cf\ \Secref{sec:flows}). Certainly, once Alice submits her application to HappyCorp, she does not need to wait for the response; she can move to her next goal while the other Flows run concurrently.

\def\thefootnote{\arabic{footnote}}

\footnotetext{For more details on meta-reasoning Flows see \Secref{sec:conclusion}}

We showcase the potential of the proposed framework and library by investigating complex collaborative and structured reasoning patterns on the challenging task of competitive coding, a mind sport involving participants trying to solve problems defined by a natural language description. 

\xhdr{Contributions} (i) We propose \frameworkName{}, a conceptual framework providing an abstraction that simplifies the design and implementation of arbitrarily nested interactions while enabling concurrency. \frameworkName{} can represent \textit{any} interaction and provides a common framework for reasoning about interaction patterns, specifying hypotheses, and structuring research more broadly. 
(ii) We open-source the \libraryName{} library, which embodies \frameworkName{}, together with FlowVerse, which is a repository of Flows that can be readily used, extended, and composed into novel, more complex Flows. (iii) We leverage \frameworkName{} and the accompanying library to systematically investigate the benefits of complex interactions for solving competitive coding problems and develop AI-only Flows adding +$21$ and human--AI Flows adding +$54$ absolute points in terms of solve rate. 


\section{Related Work}
\label{sec:related_work}

\xhdr{Existing libraries for modeling structured interactions}
LangChain \citep{langchain2022} has become the go-to library for creating applications using large language models. However, most recent works involving structured interaction, such as Cameleon \citep{Lu2023ChameleonPC}, Camel \cite{li2023camel}, HuggingGPT \cite{Shen2023HuggingGPTSA}, 
and the concurrent works MetaGPT \citep{DBLP:journals/corr/abs-2308-00352} and AutoGen \citep{DBLP:journals/corr/abs-2308-08155} all come with their own library. 
Researchers opt to implement bespoke solutions due to the lack of a general yet efficient abstraction for modeling and designing structured interactions as well as the infrastructure to implement them, that should enable and facilitate open-ended exploration of novel ideas. In this work, we develop such an abstraction, \frameworkName{}, which, in concert with \libraryName{}, fills this lacuna.

\xhdr{Impact of Flows} Crucially, the framework can implement any algorithm and efficiently covers all prior works on AI-AI, human-AI interactions, as well as prompt engineering (\cf \Appref{app:other_work}). 
These works focusing on specific Flow instantiations have demonstrated that structured interactions \textit{can} yield performance gains across tasks and models. However, recent results put the universality of previously published results into question (e.g., \citet{DBLP:journals/corr/abs-2310-01798}) and highlight the necessity for more systematic research. To support these research efforts, we develop the theoretical and practical infrastructure for modeling, implementation, and systematic study structured interactions of arbitrary complexity. We demonstrate the benefits of the proposed infrastructure by conducting experiments that thoroughly investigate multiple core interaction patterns, including Human-AI collaboration, and their combinations, while accounting for data contamination and variance in the results, both of which are, surprisingly, not currently a standard.


\textbf{Competitive coding (CC).} With the advent of transformers, \citet{li2022competition} finetuned an LLM on GitHub code repositories, and a dataset scraped from Codeforces. Recently, \citet{zelikman2022parsel} proposed decomposing CC problems into function descriptions and, for each function description, using an LLM to generate the implementation in a modular way. While these methods yield promising results, CC remains a challenging task far from being solved \citep{OpenAI2023GPT4TR}.
As such it presents itself as an ideal test bed for thoroughly studying the benefits of collaborative and structured reasoning interactions.


\section{Flows}
\label{sec:flows}

This section introduces \textit{Flows} as a conceptual framework, describes its benefits, and presents the \libraryName{} library, which embodies the framework.

\subsection{\frameworkName{} as a Conceptual Framework}

The framework is centered around 
\textit{Flows} 
and \textit{messages}. 
Flows represent the fundamental building block of computation. They are independent, self-contained, goal-driven entities able to complete a semantically meaningful unit of work.
To exchange information, Flows communicate via a standardized message-based interface. Messages can be of any type the recipient Flow can process.

We differentiate between two types of Flows: Atomic and Composite.\footnote{The concept of a Flow is sufficient for modeling any interaction. We introduce this distinction as it improves the exposition and simplifies the implementation.} Atomic Flows complete the work directly by leveraging \textit{tools}. 
Tools can be as simple as a textual sequence specifying a (simple) Flow's fixed response or as complex as a compiler, a search engine, powerful AI systems like LLaMA \cite{Touvron2023LLaMAOA, Touvron2023Llama2O}, Stable Diffusion \cite{DBLP:journals/corr/abs-2112-10752}, and GPT-4; or even a human. Notably, in the \frameworkName{} framework, AI systems correspond to tools. 
An Atomic Flow is effectively a minimal wrapper around a tool and achieves two things: (i) it fully specifies the tool (e.g., the most basic Atomic Flow around GPT-4 would specify the prompts and the generation parameters); and (ii) it abstracts the complexity of the internal computation by exposing only a standard message-based interface for exchanging information with other Flows. 
Examples of Atomic Flows include wrappers around chain-of-thought prompted GPT-4 for solving math reasoning problems, few-shot prompted LLaMA for question answering, an existing chatbot, a search engine API, or an interface with a human. 

Composite Flows accomplish more challenging, higher-level goals by leveraging and coordinating other Flows. Crucially, thanks to their local state and standardized interface, Composite Flows can readily invoke Atomic Flows or other Composite Flows as part of compositional, structured interactions of arbitrary complexity. Enabling research on effective patterns of interaction is one of the main goals of our work. General examples of such patterns include (i) factorizing the problem into simpler problems (\ie, divide and conquer); (ii) evaluating (sub-)solutions at inference time (\ie, feedback); and (iii) incorporating external information or a tool. 
Importantly, Flows can readily invoke other, potentially heavily optimized, specialized Flows to complete specific (sub-)tasks as part of an interaction, leading to complicated behavior.
One example of a Composite Flow is ReAct \cite{yao2023react}. ReAct is a sequential Flow that structures the problem-solving procedure in two steps: a Flow selects the next action out of a predefined set of actions, and another Flow executes it. The two steps are performed until an answer is obtained. Another prominent example, AutoGPT, extends the ReAct Flow with a Memory Flow and an optional Human Feedback Flow. More generally, our framework provides a unified view of prior work, which we make explicit in \Appref{app:other_work}.

Importantly, as illustrated in \Figref{fig:flows}, Composite Flows can script an arbitrarily complex pattern (i) precisely specifying an interaction (\eg, generate code, execute tests, brainstorm potential reasons for failure, etc.); or (ii) defining a high-level, meta-reasoning process in which a Flow could bring about dynamic unconstrained interactions.

\xhdr{Key properties} The proposed framework is characterized by the following key properties:
\begin{itemize}
\denselist
    \item Flows are the compositional building blocks of computation.
    \item Flows encapsulate a local, isolated state.
    \item Flows interact only via messages.
    \item Flows' behaviour depends only on their internal state and the input message.
    \item Flows can send messages to other Flows and create new Flows.
\end{itemize}

\xhdr{Connection to the \textit{Actor} model} \textit{Flows} is fundamentally a framework modeling the computation underlying interactions. 
As such, it shares key design principles with the \textit{Actor} model \cite{Hewitt1973AUM} --- a mathematical model of concurrent computation. Similarly to \textit{Flows}, in the \textit{Actor} model, an Actor is a concurrent computation entity that can communicate with other Actors exclusively through an asynchronous message-passing interface. By encapsulating the state and the computation within individual Actors, the model provides a high-level abstraction for effectively managing and reasoning about complex concurrent and distributed systems, completely avoiding issues associated with shared states, race conditions, and deadlocks. 
These benefits are similar in nature to those observed in the domain of interactions.
The main distinction between the proposed framework and the \textit{Actor} model lies in their respective communication protocols. Concretely, while the \textit{Actor} model prescribes purely asynchronous communication,  \textit{Flows} natively supports synchronous communication, which is essential for the implementation of structured reasoning. 
Interestingly, a similar deviation from the ``pure'' \textit{Actor} model can be identified in the implementation of Erlang, a concurrent programming language based on it \cite{DBLP:phd/basesearch/Armstrong03}. 
Overall, the shared design choices still make \textit{Flows} inherently concurrency\hyp friendly from the practical perspective and are sufficient for important results from the five decades of extensive studies of the \textit{Actor} model, such as the fact that every physically possible computation can be directly implemented using Actors \cite{Hewitt2010ActorMO}, to transfer to \textit{Flows}.

\subsection{Why \textit{Flows}?}
\xhdr{Modularity} \textit{Flows} introduces a higher-level abstraction that isolates the state of individual Flows and specifies message-based communication as the only interface through which Flows can interact. This ensures perfect modularity by design.

\xhdr{Reduction of complexity} The framework ensures the complexity of the computation performed by a Flow is fully abstracted behind the universal message-based interface. This enables an intuitive and simple design of arbitrarily complex interactions from basic building blocks.

\xhdr{Systematicity, flexibility, and reusability} The separation of responsibility allows for modules to be developed and studied systematically in isolation or as part of different interactions. Once the correctness and the benefits of a Flow have been established, it can be readily used in developing novel Flows or as a drop-in replacement for less effective Flows leveraged in completing similar goals.

\xhdr{Concurrency} The proposed framework's design is consistent with the Actor model, one of the most prominent models of concurrent computation. As a consequence, \textit{Flows} can readily support any setting in which Flows run concurrently.

\subsection{The \libraryName{} Library} 
Accompanying \frameworkName{}, we release the \libraryName{} library, which embodies the framework. In addition to the inherent benefits that come with the framework, the library comes with the following add-ons:
(i) FlowVerse: a repository (to which anyone can contribute) of Flows that can be readily used, extended, or composed
into novel, more complex Flows. \textit{Flows} allows for existing ``tools'' (as well as ``models'', ``chains'', ``agents'', \etc)\ to be readily incorporated by wrapping them in an Atomic Flow; 
(ii) a detailed logging infrastructure enabling transparent debugging, analysis, and research in optimizing (i.e., learning or fine-tuning) Flows.

\section{Competitive Coding Flows}
\label{sec:cc_flows}

This work investigates the potential of structured interactions for solving competitive coding (CC) problems. In CC, given a natural language description and a few input--output examples, the task is to generate code that will produce the expected output for all of the hidden input--output test cases associated with the problem. \Figref{fig:example_coding_flows} provides examples.

\begin{figure*}[t]
    \centering   
    \includegraphics[scale=1, width=0.80\textwidth]{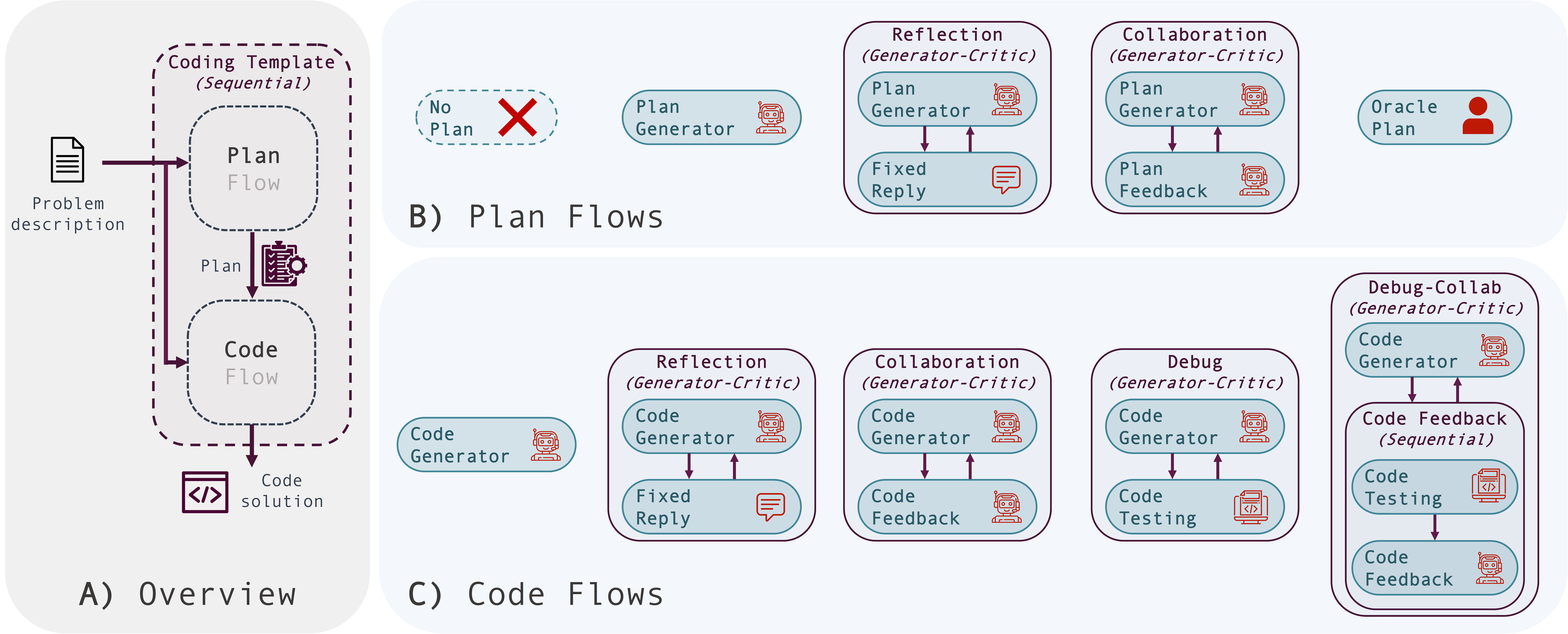}
    \caption{\textbf{Competitive coding Flows.} At the highest level, we consider planning as a specific structured reasoning pattern for problem decomposition. In particular, the Plan Flow generates a solution strategy and passes it to the Code Flow, which implements it, as depicted in A). B) and C) depict the different choices of sub-Flows used as Plan and Code Flows in the experiments. Notably, we explore the impact of human-AI collaboration at the plan level and refinement with different types of \textit{feedback}: i) fixed reply encouraging reflection; ii) AI generated feedback; iii) code testing results as feedback; iv) AI generated feedback grounded in code testing results.}
    \label{fig:coding_flows}
\end{figure*}

We focus the analysis on three canonical dimensions of interactions:
(i) problem decomposition as structured reasoning;
(ii) human-AI collaboration; and 
(iii) refinement with various feedback types.
By providing a common language for clearly specifying interactions as well as the capability to flexibly compose, exchange, and extend them, the framework makes it possible to study the space of complex interactions in a principled fashion. In the rest of the section, we describe the specific Flows used in the experiments, depicted in \Figref{fig:coding_flows}.

\xhdr{Problem decomposition}
Planning has been an integral intermediate step in recent work \citep{Lu2023ChameleonPC, Shen2023HuggingGPTSA, yao2023react}. Similar decomposition is natural in the context of CC as well. In particular, we approach the task in two steps: generating a solution strategy by a Plan Flow and then generating the corresponding code by a Code Flow. This is depicted by panel A in \Figref{fig:coding_flows}.

\xhdr{Human-AI collaboration} When designing human-AI collaborations, it is essential to take the costs of human interaction into account \cite{DBLP:conf/chi/Horvitz99, DBLP:conf/chi/AmershiWVFNCSIB19,DBLP:journals/corr/abs-2306-04930}. By providing immense flexibility, \textit{Flows} can support research in the design of interactions involving humans as computational building blocks in a way that maximizes the utility of the overall computation with a minimal human effort.
In the context of CC, we hypothesize that a human can be effectively incorporated at the plan level to provide a short ``oracle'' plan in natural language. We operationalize this by an (Atomic) Human Flow, illustrated in Panel B of \Figref{fig:coding_flows} as the \textit{Oracle Plan} Flow.

\xhdr{Refinement with various feedback types} 
Iterative refinement is a general problem-solving strategy successfully deployed across various disciplines \citep{Perrakis1999AutomatedPM, Reid2022LearningTM, Schick2022PEERAC, Saharia2021ImageSV}. The strategy revolves around the idea that a solution can be gradually improved through a mechanism for analysis, modification, and re-evaluation. The design of this ``\textit{feedback}'' mechanism is critical for the effectiveness of the problem-solving strategy. The conceptual framework, paired with the accompanying library, provides the infrastructure to support the design, implementation, and principled research of effective refinement strategies and feedback mechanisms. In this work, we consider a canonical iterative refinement setup where a \textit{generator} Flow is tasked with generating the solution, and a \textit{critic} Flow provides feedback on the proposed solution. 
We consider two feedback types in the context of both the Plan and the Code Flow: (i) Reflection Flow: the feedback consists of a fixed message encouraging the model to reflect on important aspects of the proposed solution; (ii) Collaboration Flow: the feedback is provided by an AI system that ``evaluates'' the proposed solution. Furthermore, we explore two more code-specific feedback types: 
(i) Debug Flow: the feedback message corresponds to the results from executing the code and testing it against the examples provided in the problem description; (ii) Debug--Collab Flow: the feedback is provided by an AI system with access to the code testing results, effectively, grounding the feedback and allowing more systematic reasoning about the potential causes of failure.

We refer to Flows using the following convention: \textit{CodeFlowName} when no plan is generated and \textit{PlanFlowName-CodeFlowName} otherwise.

\section{Experimental Setup}

\textbf{Data.} 
We scrape publicly available problems from one of the most popular websites hosting CC contests, Codeforces \cite{codeforces}, and LeetCode \cite{leetcode}, which cover a broad spectrum of problems ranging from easy interview questions to hard CC problems (see \Appref{app:data} for more details).
The datasets cover problems from $2020$-August-$21$ to $2023$-March-$26$ for CodeForces, and from $2013$-October-$25$ to $2023$-April-$09$ for LeetCode. 
Importantly, to study the effect of structured interactions (\ie, different Flows) in a principled manner, it is crucial to account for the possibility of \textit{data contamination}, \ie, that some of the test data has been seen during training \cite{DBLP:conf/acl/Magar022}. Containing problems published over an extended period up to a few months ago (at the time of writing), our datasets allow for reliable identification of the training data cutoff date that can help with addressing this issue. 
Prior code evaluation datasets like APPS  \citep{hendrycksapps2021}, HumanEval \citep{Chen2021EvaluatingLL}, and CodeContests \citep{li2022competition} lack problem release dates, and considering the lack of publicly available information about LLMs' training data, can likely lead to confounded evaluation of models' memorization and generalization abilities.

\textbf{Code testing and solution evaluation.}
Just like a human participant, the Debug Flow has access only to the input--output example pairs contained in the problem description and, at inference time, uses a local code testing infrastructure to evaluate (intermediate) solution candidates. Crucially, these examples cover only a few simple cases, and generating outputs consistent with them does not imply the code corresponds to a correct solution. A solution is considered correct if it passes all the hidden test cases. To determine correctness, we leverage online evaluators that submit candidate solutions to the websites' online judges, ensuring authoritative results. For many of the Codeforces problems, we also support local evaluation based on a comprehensive set of hidden test cases we managed to scrape. For more details, see \Appref{app:code_testing}.

\textbf{Models and Flows.}
We experiment with the competitive coding Flows described in \Secref{sec:cc_flows}, and GPT-4 \citep{OpenAI2023GPT4TR} as the LLM tool of choice. See \Appref{app:prompting} for the specific prompts. Also, the code to reproduce the experiments in the paper is available in the project's GitHub repository.

\textbf{Evaluation metrics.} The most common evaluation metric for code generation is pass@$k$, corresponding to the probability that in a set of $k$ sampled candidates, there will be at least one correct solution \cite{Chen2021EvaluatingLL}. To better align with practical use cases, we focus on pass@1, \ie the solve rate when averaged across the problem set. We report a point estimate and a 95\% confidence interval constructed from 1000 bootstrap resamples.

\textbf{Compute and cost.}
All the experiments, including the most complex Flows, can be performed on commodity hardware relatively cheaply. For instance, the costs associated with querying the OpenAI API for generating \Tabref{tab:main} amount to \$1000. 

\section{Experimental Results}\label{sec:results}

We first study the generalization ability of representative Flows and empirically identify GPT-4's knowledge-cutoff date. Next, we perform a focused analysis along the dimensions described in \Secref{sec:cc_flows}.

\begin{table*}[ht]
\caption{\textbf{Main Results.} Performance of competitive coding Flows on Codeforces and LeetCode, with direct inference (Code) as baseline.}
\vskip 0.15in
\label{tab:main}
\centering
\small
\scalebox{0.92}{
\begin{tabular}{@{}lc|c||c|c|c|c|c|c@{}}
\toprule
 & \multicolumn{2}{c||}{{\textbf{Codeforces}}} & \multicolumn{6}{c}{{\textbf{Leetcode}}} \\
 & \multicolumn{1}{c|}{Pre-cutoff} & \multicolumn{1}{c||}{Post-cutoff} & \multicolumn{3}{c|}{Pre-cutoff} & \multicolumn{3}{c}{Post-cutoff} \\
 &  &  & Easy & Medium & Hard & Easy & Medium & Hard \\
\midrule
\hspace{4mm} \colorbox{yellow!20}{Code} & 71.8 {\scriptsize±11.0} & 26.9 {\scriptsize±11.0} & 97.8 {\scriptsize±3.1} & 93.4 {\scriptsize±5.4} & 66.7 {\scriptsize±10.9} & 76.3 {\scriptsize±8.6} & 25.1 {\scriptsize±8.9} & 8.0 {\scriptsize±5.5} \\
\hspace{4mm} \colorbox{yellow!20}{Code\_Reflection} & +9.3 {\scriptsize±9.7} & +0.0 {\scriptsize±10.6} & +0.0 {\scriptsize±3.1} & +0.0 {\scriptsize±5.4} & +1.2 {\scriptsize±10.6} & +0.9 {\scriptsize±8.1} & +5.4 {\scriptsize±9.4} & +3.5 {\scriptsize±6.6} \\
\hspace{4mm} \colorbox{yellow!20}{Code\_Collaboration} & +4.8 {\scriptsize±10.5} & +9.6 {\scriptsize±11.8} & +0.0 {\scriptsize±3.1} & -2.3 {\scriptsize±6.0} & -0.1 {\scriptsize±10.9} & -3.2 {\scriptsize±8.7} & +0.0 {\scriptsize±8.7} & +1.2 {\scriptsize±5.9} \\
\hspace{4mm} \colorbox{yellow!20}{Code\_Debug} & +12.7 {\scriptsize±8.6} & +7.9 {\scriptsize±11.6} & +0.0 {\scriptsize±3.1} & +1.1 {\scriptsize±5.0} & +6.9 {\scriptsize±10.0} & +7.7 {\scriptsize±7.3} & +7.7 {\scriptsize±9.6} & +2.4 {\scriptsize±6.3} \\
\hspace{4mm} \colorbox{yellow!20}{Code\_Debug\_Collab} & +12.6 {\scriptsize±8.9} & +20.6 {\scriptsize±12.1} & +0.0 {\scriptsize±3.1} & +0.0 {\scriptsize±5.4} & +5.5 {\scriptsize±10.4} & +7.5 {\scriptsize±7.4} & +9.8 {\scriptsize±9.7} & +1.2 {\scriptsize±6.0} \\
\midrule
\hspace{4mm} \colorbox{violet!10}{Plan-Code} & -1.6 {\scriptsize±11.0} & +8.0 {\scriptsize±11.6} & -3.1 {\scriptsize±4.5} & -2.3 {\scriptsize±5.9} & -9.7 {\scriptsize±11.2} & +2.3 {\scriptsize±8.3} & +3.2 {\scriptsize±9.1} & -3.4 {\scriptsize±4.3} \\
\hspace{4mm} \colorbox{violet!10}{Plan\_Reflection-Code} & -3.3 {\scriptsize±11.6} & +4.8 {\scriptsize±11.6} & -2.1 {\scriptsize±4.1} & -4.5 {\scriptsize±6.6} & -3.1 {\scriptsize±10.7} & +1.2 {\scriptsize±8.3} & -3.3 {\scriptsize±8.5} & +0.0 {\scriptsize±5.5} \\
\hspace{4mm} \colorbox{violet!10}{Plan\_Collaboration-Code} & -4.8 {\scriptsize±11.5} & +6.3 {\scriptsize±11.4} & -1.1 {\scriptsize±3.7} & -2.3 {\scriptsize±6.1} & -7.2 {\scriptsize±11.2} & -2.0 {\scriptsize±8.6} & +0.1 {\scriptsize±9.0} & +1.2 {\scriptsize±5.8} \\
\midrule
\hspace{4mm} \colorbox{violet!10}{Plan\_Oracle-Code} & +11.0 {\scriptsize±9.4} & +47.6 {\scriptsize±10.7} & -- & -- & -- & -- & -- & -- \\
\hspace{4mm} \colorbox{violet!10}{\begin{tabular}{@{}l}Plan\_Oracle-Code\_\\Debug\_Collab\end{tabular}} & +23.0 {\scriptsize±5.2} & +53.9 {\scriptsize±9.5} & -- & -- & -- & -- & -- & -- \\
\bottomrule
\end{tabular}}
\end{table*}

\subsection{Performance of Coding Flows on Pre- vs.\ Post-Knowledge-Cutoff-Date Data}\label{sec:6.1} 

\begin{figure}[h]
    \centering   
    \includegraphics[scale=1, width=1.01\columnwidth]{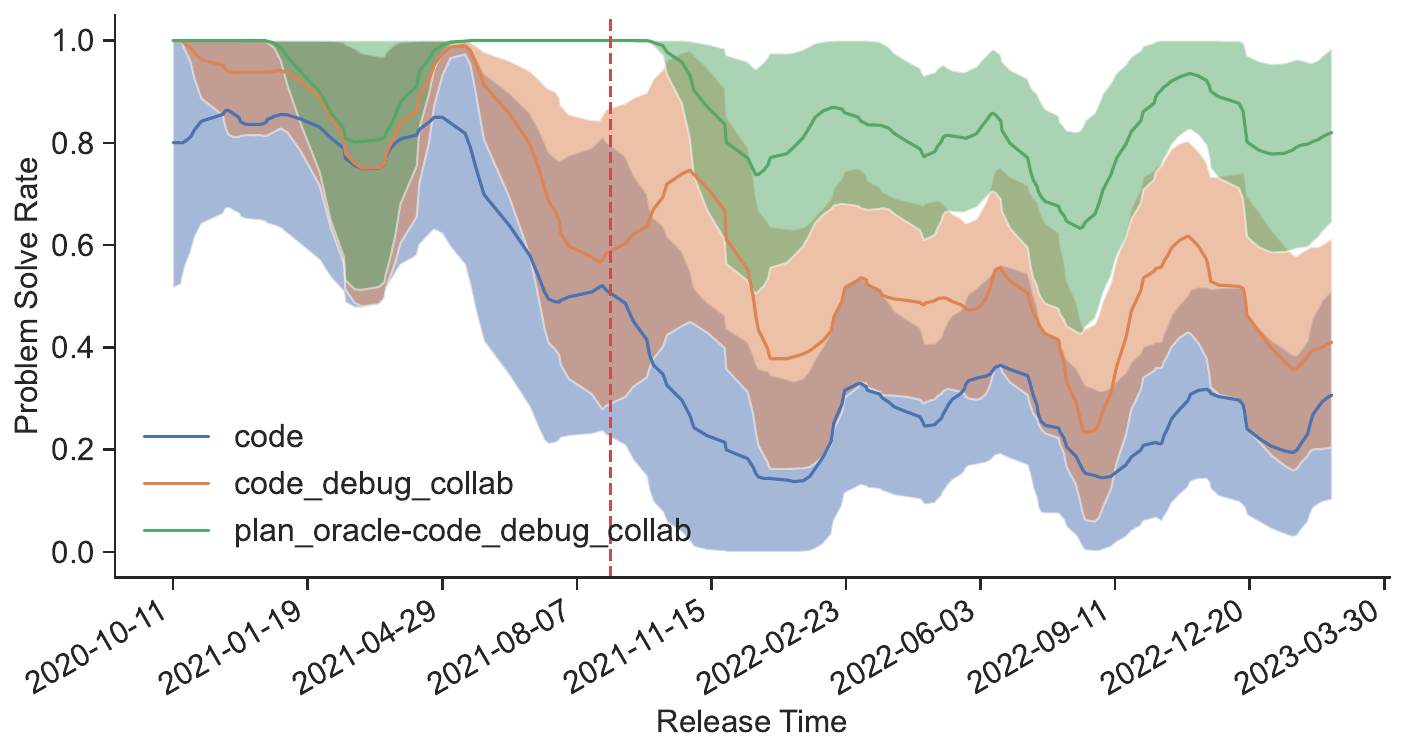}
    \caption{\textbf{Temporal analysis.} Performance is averaged over a sliding window of two months. The substantial drop in performance around the reported knowledge cutoff date for GPT-3/4 (the crimson vertical line) reveals limited generalization ability that can be alleviated through structured interactions.}
    \label{fig:temporal_analysis}
\end{figure}

In this experiment, we consider three representative Flows: (i) Code: the simplest Code Generator Flow corresponding to a single GPT-4 API call; (ii) \mbox{Code\_Debug\_Collab}: the most complex code Flow; (iii) \mbox{Plan\_Oracle-Code\_Debug\_Collab}: the most complex code Flow with human guidance at the plan level. We perform the analysis by running the three Flows on Codeforces problems released from October 2020 to April 2023 and averaging the performance over a sliding window of two months. The results are reported in \Figref{fig:temporal_analysis}.

We observe a substantial drop in performance centered around September 2021, consistent with the knowledge cutoff date reported by OpenAI, and denote it by a vertical line on the plot. With Codeforces problems appearing in contexts outside of the contest itself (\eg, editorials), it is reasonable to assume the model has been exposed to older problems more frequently during training. This would explain why the drop spans multiple months, from May 2021 to November 2021, depending on when which data was published and crawled. 

Notably, there is a stark difference in the performance of the Code Flow on problems published before and after the knowledge cutoff data, with the solve rate decreasing from around 80\% to $23\%$. While still experiencing a substantial performance drop, the \mbox{Code\_Debug\_Collab} Flow doubles the solve rate on novel problems to around 45\%. Provided with human input at the plan level, the same Flow reaches $85\%$. Overall, this highlights that GPT-4 performs poorly on novel complex reasoning problems, but structured interactions have the potential to enhance its generalization capabilities. As both GPT-4 (\ie, the Code Flow) and the more complex interactions (Flows) exhibit qualitatively different behavior on novel data, to draw accurate conclusions, it is critical that data contamination is taken into serious consideration when designing experiments and interpreting results.

\subsection{Comparing Competitive Coding Flows}

Table \ref{tab:main} reports the performance of the systematically chosen set of Flows described in \Secref{sec:cc_flows}. Rows $6$--$10$ correspond to Flows comprising \colorbox{violet!10}{planning and coding}, while rows $1$--$5$ perform the  \colorbox{yellow!20}{coding directly}. 
In line with the findings of the previous section, we separately consider the performance on problems published before and after the knowledge cutoff date of September 2021.

\xhdr{Problem decomposition}
The idea behind planning before implementing the solution is to decouple the high-level reasoning from the code implementation. To analyze the effectiveness of this pattern, we compare the Code and the Plan-Code Flow.
Looking at the point estimates, in the pre-cutoff problems, introducing the plan Flow leads to decreased performance (-$1.6$ for Codeforces and -$3.1$/$2.3$/-$9.7$ for LeetCode easy/medium/hard). However, in the post-cutoff problems, incorporating a plan Flow leads to gains for Codeforces (+$8$) and LeetCode easy and medium (+$2.3$ and +$3.2$). 
While these trends are consistent, considering the confidence intervals, we see that they are not statistically significant. 
Crucially, these results do not imply that this specific problem decomposition is not valuable as it creates a lot of potential in designing an effective human-AI collaboration.

\xhdr{Human-AI collaboration}
\label{oracle}
After every contest, the Codeforces community publishes an editorial that, in addition to the code implementation, provides a short natural language description of the solution.
To simulate a Flow where a human provides high-level guidance at the core of the reasoning process, we scrape the solution descriptions and pass them as human-generated plans.  
The results are striking: despite being only a few sentences long, human-provided plans lead to a substantial performance increase (from $26.9\%$ to $74.5\%$ and from $47.5\%$ to $80.8\%$ on novel problems, when the code is generated by \mbox{Code} and \mbox{Code\_Debug\_Collab Flows}, respectively).
First and foremost, these results showcase the opportunities created by \frameworkName{} for designing, implementing, and studying Human-AI collaboration as a key component of structured interactions. 
Second, specific to the problem of competitive coding, they validate the hypothesis that high-quality plans are important, suggesting that the design of more effective plan Flows is a promising direction to explore in the future. 
Last but not least, the results highlight the necessity of more systematic research, as patterns seemingly not valuable in one Flow, such as the simple plan-code structured reasoning problem decomposition, can provide immense value as part of another Flow.

\xhdr{Refinement with various feedback types} 
We find that \mbox{Code\_Reflection} and \mbox{Code\_Collaboration} lead to limited improvements among the code Flows. The two exceptions are Codeforces pre-cutoff ($+9.3$) for the former and Codeforces post-cutoff ($+9.6$) for the latter pattern. While close, these results are not statistically significant. On the other hand, the Flows providing grounded feedback, \mbox{Code\_Debug} and \mbox{Code\_Debug\_Collab}, lead to consistent and statistically significant improvements, most notable on the novel Codeforces problems where performance increases from $26.9$, without feedback, to $47.5$, when the refinement is based on AI-generated feedback grounded in tests. On LeetCode, these improvements are smaller in magnitude. We suspect this is a consequence of the examples provided with the problem description being more simple than those in Codeforces, leading to false positives and, thereby, incorrect grounding, affecting the feedback quality. 
This could be addressed by generating additional tests with a \mbox{Test\_Case\_Generator} Flow, a direction we leave for future work to explore. Finally, in the plan Flows, where we consider Reflection and Collaboration (without grounding), we find that refinement does not provide statistically significant benefits.

\xhdrNoPeriod{Overall}, our findings provide several important insights:
(i) the direct benefit of problem decomposition hinges on the quality of the intermediate steps;
(ii) involving humans at the core high-level reasoning process yields major improvements as humans can easily provide high-quality, grounded feedback;
(iii) strategic problem decomposition is a powerful strategy for creating opportunities for effective Human--AI collaboration;  
(iv) the effectiveness of refinement patterns is not universal and depends on the quality of the starting solution and the feedback (\eg, the level of grounding), and the model's ability to incorporate that feedback modulated through the feedback's specificity and the model's capabilities. 
This analysis paints a more complex picture than what is reported by prior work for simple interactions.

\FloatBarrier

\section{Discussion}
\label{sec:conclusion}

\xhdr{Simplicity and systematicity}
Thanks to its key properties, \frameworkName{}, together with \libraryName{}, provides an infrastructure that greatly simplifies the design and implementation of open-ended interactions, with a capability to flexibly isolate, compose, replace, or modify sub-Flows. The experiments demonstrate that carefully designed interactions can substantially improve generalization.
However, they also reveal that the effectiveness of particular interaction patterns is not universal; instead, there are many factors at play. 
As researchers, we need to clearly specify the patterns we are studying, clearly communicate our hypotheses, and study them both in isolation and as parts of other interactions across different datasets or/and tasks. 
Furthermore, it is critical that data contamination is taken into serious consideration when designing experiments and drawing conclusions, and error bars become a standard in the field.

\xhdr{Cost and performance Optimization} In our experiments, we used ``off-the-shelf'' LLMs that have not been specifically optimized for collaboration. Performance (and compute costs) can be substantially improved by fine-tuning models to collaborate more effectively, generally or toward specialized roles (e.g., controller or critic). Learning requires data, and to support research in this direction, \libraryName{} implements detailed logging mechanisms of Flow runs.

\xhdr{Meta-reasoning Flows and asynchronous execution} 
Cognitive science research in metacognition and meta-reasoning suggests the existence of meta-level monitoring and control processes underlying cognition \cite{ACKERMAN2017607}. 
Since \frameworkName{} supports asynchronous execution of sub-Flows, it makes it possible to achieve similar asynchronous meta-cognition for autonomous AI systems moving beyond a single LLM call serving as a controller \cite{babyagi2023, autogpt2023}. 
For example, distributed and asynchronous execution of Flows such as FunSearch \cite{romera2023mathematical} is naturally supported by \frameworkName{}.

\section{Conclusion}
In this paper, we propose \frameworkName{}, an abstraction that, in concert with the accompanying library \libraryName{}, provides the theoretical and practical infrastructure with a modular and concurrency-friendly design, which enables and facilitates the modeling, implementation, and systematic study of arbitrarily complex structured interactions. 
We thoroughly investigate multiple core interaction patterns, including Human-AI collaboration, and their combinations, while accounting for data contamination and the variance in the results. The investigation shows that the developed AI-only Flows add +21 and human--AI Flows add +54 absolute points in terms of solve rate, and highlights the effect of data contamination, variance, and non-universality of results. 
Overall, our experiments establish the potential of Flows, the necessity of more systematic research, and the value brought by \frameworkName{} and \libraryName{} in support of these research efforts.    
On the one hand, \textit{Flows} provides a high-level abstraction enabling the design and implementation of interactions of arbitrary complexity. On the other, it offers a common framework for reasoning about interaction patterns, specifying hypotheses, and structuring research. We hope the framework will serve as a solid basis for practical and theoretical innovations, paving the way toward ever more useful AI, similar to the Actor model's role for concurrent and distributed systems.

\bibliography{00_main}
\bibliographystyle{icml2024}

\newpage
\appendix
\onecolumn

\clearpage
\section{Appendix}
\subsection{Data}
\label{app:data}
\begin{figure*}[h]
    \centering   
    \includegraphics[scale=1, width=0.58\paperwidth]{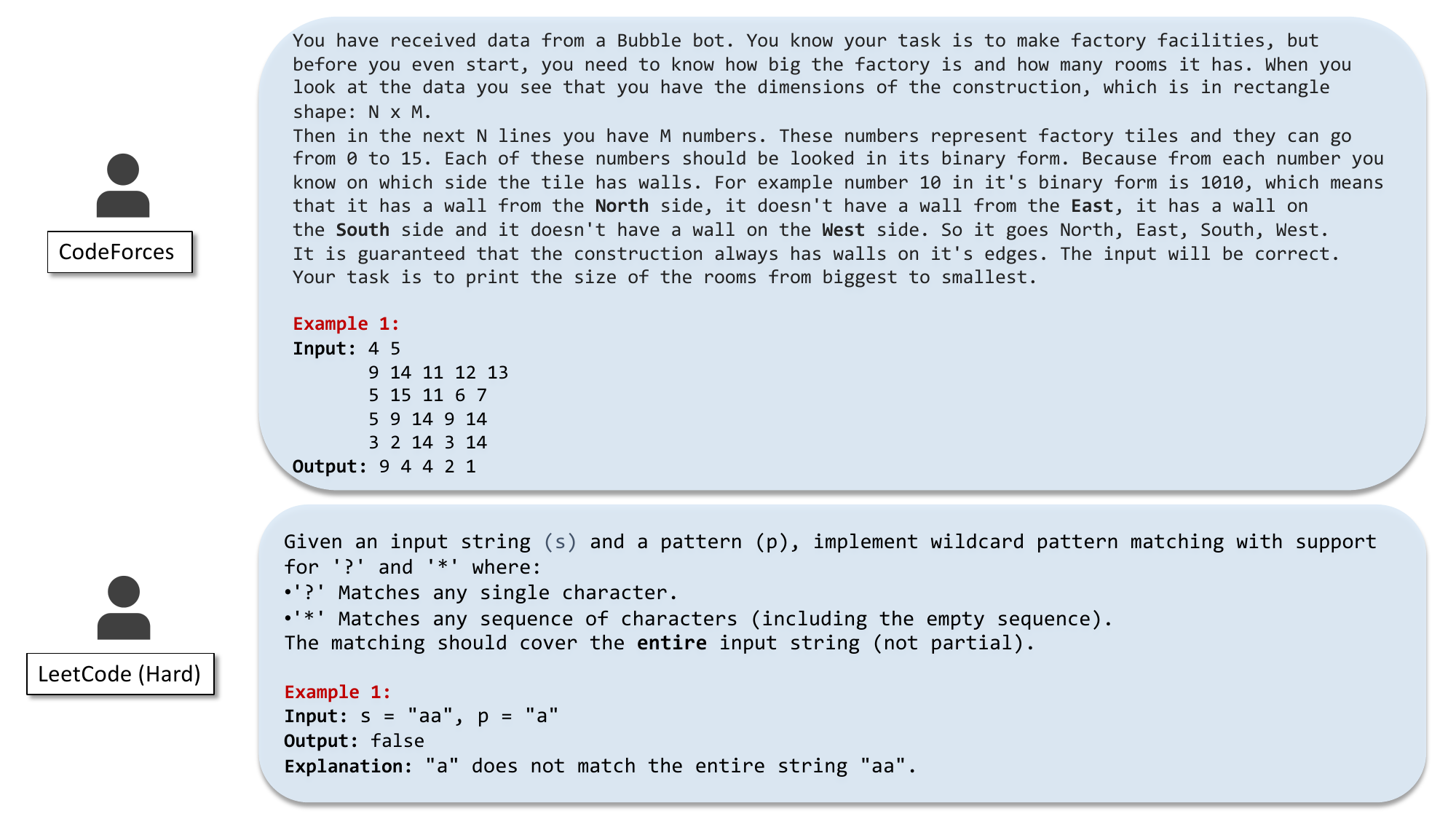}
    \caption{\textbf{Examples of competitive coding problems from Codeforces and LeetCode.}}
    \label{fig:example_coding_flows}
\end{figure*}

Example Codeforces and LeetCode problems are provided in \Figref{fig:example_coding_flows}. 

In the first experiment, the temporal analysis, we use 239 Codeforces problems ranging from October 2020 to April 2023. In the second experiment, we have 136 problems for Codeforces (some problems are dropped in order to keep the pre-cutoff and post-cutoff buckets equal to 68) and 558 problems for LeetCode (93 for each of the six buckets). 
Additionally, to support research in the area, we set up an AI competitive coding challenge based on a dataset of Codeforces problems of various difficulties published after the knowledge cutoff date. More details about the CC competition are available in \Appref{app:cc-competition}.

\subsection{Code Testing and Solution Evaluation}
\label{app:code_testing}

The solution evaluation requires a set of input--output pairs, hidden from the user, that comprehensively test the behavior of the program. To compute the final results, we have implemented an online evaluation infrastructure that submits the candidate solutions to the websites' online judges and automatically scrapes the judgment. This mechanism ensures authoritative results. 

For many of the Codeforces problems, we managed to scrape (sometimes a subset) of the hidden tests, allowing us to use a faster, local infrastructure for evaluating candidate solutions. On the other hand, LeetCode does not expose any of the hidden tests publicly.

For code testing at inference time, just like a human would, we rely on tests constructed from the (public) input--output example pairs contained in the problem description.

\subsection{Concurrent and Previous Works as Specific Instances of Flows}\label{app:other_work}
The introduction of LLMs such as BARD, GPT-3, ChatGPT, and its latest version, GPT-4, has led to a breakthrough in AI.
This has enabled many exciting developments like CoT, HuggingGPT, AutoGPT, AgentGPT, and BabyAGI.
In this section, we demonstrate how \textit{Flows} provides a unified view encompassing concurrent and previous work as specific Flow instances. 
The details are provided in Figure \ref{fig:previous_works} and Table. \ref{tab:previous_flows}.

    \begin{enumerate}
        \item \textbf{Few shot Prompting} (FS) \citep{NEURIPS2020_1457c0d6} consists in providing a few input-output examples within the prompt, acting as demonstrations to enable the LLM to perform a specific task. This technique relies on the LLM's emergent in-context learning ability to extrapolate from these limited examples and infer how to solve the task in general.
        
        \item \textbf{Chain of Thoughts} (CoT) \citep{wei2022chain} is a prompting method (atomic Flow) that allows LLMs to generate a series of intermediate natural language reasoning steps that lead to the final output.

        \item \textbf{Tree of Thoughts} (ToT) \citep{Yao2023TreeOT} is a framework that enables (\textit{orchestration}) exploration over coherent units of text (thoughts) that serve as intermediate steps toward problem-solving. ToT allows LLMs to perform deliberate decision-making by considering multiple different reasoning paths and self-evaluating choices to decide the next course of action, as well as looking ahead or backtracking when necessary to make global choices.  

        \item \textbf{Program of Thoughts} (PoT) \citep{Chen2022ProgramOT} is a prompting method that allows language models (mainly Codex) to express the reasoning process as a program. The computation is relegated to an external program, which executes the generated programs to derive the answer. 

        \item \textbf{Mutimodal CoT} (M-CoT) \citep{Zhang2023MultimodalCR} is a method that incorporates language (text) and vision (images) modalities into a two-stage framework that separates rationale generation and answer inference. To facilitate the interaction between modalities in M-CoT, smaller language models (LMs) are fine-tuned by fusing multimodal features.

        \item \textbf{ToolFormer} \citep{Schick2023ToolformerLM} is a model that is trained to decide which APIs to call, when to call them, what arguments to pass, and how to incorporate the results into future tokens prediction. 
        
        \item \textbf{ReAct} \cite{yao2023react} is a framework that uses LLMs to generate reasoning traces and task-specific actions sequentially. The framework allows for greater synergy between the two: reasoning traces help the model induce, track, and update action plans and handle exceptions, while actions allow it to interface with external sources, such as knowledge bases or environments, to gather additional information. 
\begin{figure*}[t]
    \centering   
    \includegraphics[scale=1, width=0.64\paperwidth]{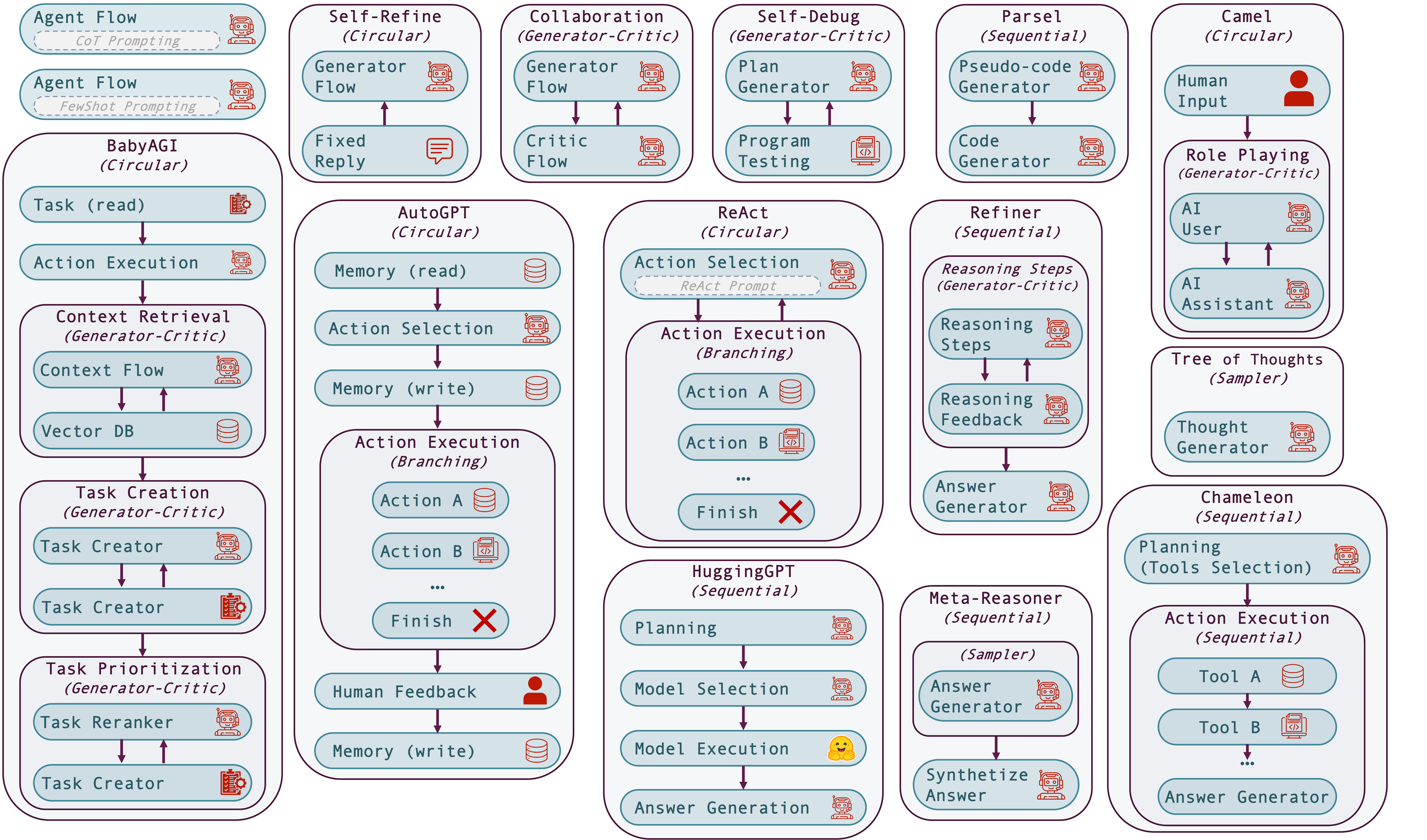}
    \caption{\textbf{Previous works are specific Flows.} We depict a selected subset of previous works incorporating structured reasoning and/or interactions between AI agents, tools, and humans, through the lens of the Flows framework. This demonstrates that Flows is a powerful language for describing, conceptualizing, and disseminating structured interaction patterns.}
    \label{fig:previous_works}
\end{figure*}
        \item \textbf{Parsel} \citep{zelikman2022parsel} is a framework that enables the automatic implementation and validation of complex algorithms with code LLMs. The framework first synthesizes an intermediate representation based on the Parsel language and can then apply a variety of postprocessing tools. Code is generated in a next step.
        
        \item \textbf{REFINER} \citep{paul2023refiner} is a framework for LMs to explicitly generate intermediate reasoning steps while interacting with a critic model that provides automated feedback on the reasoning. 
        
        \item \textbf{Self-Refine} \citep{madaan2023selfrefine} is a framework for LLMs to generate coherent outputs. The main idea is that an LLM will initially generate an output while the same LLM provides feedback for its output and uses it to refine itself iteratively.
        
        \item \textbf{Recursively Criticize and Improve} (RCI) \citep{Kim2023LanguageMC} showed that a pre-trained large language model (LLM) agent could execute computer tasks guided by natural language using a simple prompting scheme where the agent Recursively Criticizes and Improves its output (RCI). Unlike Self-refine, this method uses two separate LLMs (ChatGPT), one for performing the task and another for criticizing. 
        
        \item \textbf{Self-Correct} \citep{welleck2023generating} is a framework that decouples a flawed base generator (an LLM) from a separate corrector that learns to iteratively correct imperfect generations. The imperfect base generator can be an off-the-self LLM or a supervised model, and the corrector model is trained. 
        
        \item \textbf{Self-Debug} \citep{Chen2023TeachingLL} is a framework that relies on external tools (SQL application or Python interpreter) to help large language models revise and debug SQL commands or Python code with bugs.

        \item \textbf{Reflexion} \citep{shinn2023reflexion} is a framework that provides a free-form reflection on whether a step was executed by LLM correctly or not and potential improvements. Unlike self-refine and self-debug, Reflexion builds a persisting memory of self-reflective experiences, which enables an agent to identify its own errors and self-suggest lessons to learn from its mistakes over time. 
        
        \item \textbf{Meta-Reasoner} \citep{Yoran2023AnsweringQB} is an approach which prompts large language models to meta-reason over multiple chains of thought rather than aggregating their answers. This approach included two steps: (i) ask LLM to generate multiple reasoning chains, (ii) ask another LLM (meta-reasoner) to reason over the multiple reasoning chains to arrive at the correct answer.

        \item \textbf{HuggingGPT} \citep{Shen2023HuggingGPTSA} is a framework that leverages LLMs (e.g., ChatGPT) to connect various AI models in machine learning communities (e.g., Hugging Face) to solve numerous sophisticated AI tasks in different modalities (such as language, vision, speech) and domains.
        
        \item \textbf{Camel} \citep{li2023camel} is a communicative agent framework involving inception prompting to guide chat agents toward task completion while maintaining consistency with human intentions.
        
        \item \textbf{Chameleon} \citep{Lu2023ChameleonPC} is a plug-and-play compositional reasoning framework that augments external tools with LLMs in a plug-and-play manner. The core idea is that an LLM-based planner assembles a sequence of tools to execute to generate the final response. The assumption is that this will be less error-prone, easily expandable to new modules, and user-friendly. 
        
        \item \textbf{AutoGPT} \citep{autogpt2023} is an experimental open-source application that leverages the capabilities of large language models (LLMs) and Chatbots such as OpenAI’s GPT-4 and Chat-GPT to create fully autonomous and customizable AI agents. It has internet access, long-term and short-term memory management.

        \item \textbf{BabyAGI} \citep{babyagi2023} is an intelligent agent capable of generating and attempting to execute tasks based on a given objective. BabyAGI operates based on three LLM flows: Task creation flow, Task prioritization flow, and Execution flow.
\end{enumerate}

\begin{table*}[t!]
  \small
  \centering
  \scalebox{0.9}{
  \begin{tabular}{@{}lccccccccc@{}}
    \toprule
    {\bf Flows} & {\bf Flow Type} & \multicolumn{4}{c}{\bf Interactions} & \multicolumn{2}{c}{\bf  Reasoning Patterns} & \textbf{Feedback} & {\bf Learning} \\
    \cmidrule(lr){3-6}\cmidrule(lr){7-8}
    {} & {} & Self &  Multi-Ag. & Human & Tools & Struct. & Plan &  & {} \\
    \midrule

    FS \citep{NEURIPS2020_1457c0d6} & Atomic & \xmark & \xmark & \xmark& \xmark & \xmark & \xmark & \xmark & \xmark \\ 
    
    CoT \citep{wei2022chain} & Atomic & \xmark & \xmark & \xmark & \xmark & $\checkmark$ & \xmark & \xmark & \xmark \\

    ToT \citep{Yao2023TreeOT} & Circular & $\checkmark$ & \xmark & \xmark & $\checkmark$  & $\checkmark$ & \xmark & \xmark & \xmark \\ 

    PoT \citep{Chen2022ProgramOT} & Seq. & \xmark & \xmark & \xmark & $\checkmark$ & $\checkmark$ & \xmark & \xmark & \xmark \\

    M-CoT \citep{Zhang2023MultimodalCR} & Seq. & \xmark & \xmark & \xmark & \xmark & $\checkmark$  & \xmark & \xmark & $\checkmark$  \\

    ToolFormer \citep{wei2022chain} & Seq. & \xmark & \xmark & \xmark & $\checkmark$ & $\checkmark$ & \xmark & \xmark & $\checkmark$ \\ 

    ReAct \citep{yao2023react} & Circular & \xmark & \xmark & \xmark & $\checkmark$ & $\checkmark$ & \xmark & \xmark & \xmark \\ 

    Parsel \citep{zelikman2022parsel} & Seq. & \xmark &  $\checkmark$ & \xmark & $\checkmark$ &  $\checkmark$ &  $\checkmark$ & \xmark & \xmark \\ 

    REFINER \citep{paul2023refiner} & Gen-Crit & \xmark & $\checkmark$  & $\checkmark$  & \xmark & $\checkmark$ & \xmark & $\checkmark$ & $\checkmark$ \\ 

    Self-Refine \citep{madaan2023selfrefine} & Gen-Crit & $\checkmark$ & \xmark & \xmark & \xmark & $\checkmark$ & \xmark & $\checkmark$ & \xmark \\ 

    RCI \citep{Kim2023LanguageMC} &  Gen-Crit & $\checkmark$ & \xmark & \xmark & $\checkmark$ & $\checkmark$ & \xmark & $\checkmark$ & \xmark \\

    Self-Correct \citep{welleck2023generating} &  Gen-Crit & $\checkmark$ & \xmark & \xmark & $\checkmark$ & $\checkmark$ & \xmark & $\checkmark$ & \xmark \\

    Self-Debug \citep{Chen2023TeachingLL} & Gen-Crit & $\checkmark$ & \xmark & \xmark & $\checkmark$ & $\checkmark$ & \xmark & $\checkmark$ & \xmark \\ 

    Reflexion \citep{shinn2023reflexion} & Gen-Crit & $\checkmark$ & \xmark & \xmark & $\checkmark$ & \xmark & \xmark & $\checkmark$ & \xmark \\ 
    
    Meta-Reasoner \citep{Yoran2023AnsweringQB} & Seq. & $\checkmark$ & $\checkmark$ & \xmark & \xmark & $\checkmark$ & \xmark & \xmark & \xmark \\

    HuggingGPT \citep{Shen2023HuggingGPTSA} & Seq. & \xmark & $\checkmark$ & \xmark & $\checkmark$ & $\checkmark$ & $\checkmark$ & \xmark & \xmark \\

    Camel  \citep{li2023camel}  & Circular & \xmark & $\checkmark$ & $\checkmark$ & \xmark & $\checkmark$ & \xmark & $\checkmark$ & \xmark \\

    Chameleon \citep{Lu2023ChameleonPC}  & Seq. & \xmark & $\checkmark$ & \xmark & $\checkmark$ & $\checkmark$ & $\checkmark$ & \xmark & \xmark \\

    AutoGPT \citep{autogpt2023}  & Circular & $\checkmark$ & $\checkmark$ & \xmark  & $\checkmark$ & $\checkmark$ & $\checkmark$ & $\checkmark$ & \xmark \\

    BabyAGI \citep{babyagi2023}  & Circular & \xmark & $\checkmark$ & \xmark & $\checkmark$ & $\checkmark$ & $\checkmark$ & \xmark & \xmark \\ 
    \bottomrule\\
  \end{tabular}}
  \caption{\textbf{Previous work.} We compare previous work across relevant dimensions.}
\label{tab:previous_flows}

\end{table*}

\subsection{Prompting} \label{app:prompting}
We provide the prompts used to obtain the results in Section \ref{sec:results}. 
Our evaluation is made possible thanks to the modular and compositional nature of \frameworkName{}. Some of the experimental setups are deeply nested, and in cases where Flows build on each other, we avoid repetition.
Note that the project's GitHub repository provides the code and data to reproduce all of the experiments in the paper. 

Direct prompting for a solution is shown in Listing \ref{listing:cf_code}. To add reflection, we use a Generator-Critic Flow to combine the code generation with a fixed reply, as shown in Listing \ref{listing:fixed_reply_reflect}.
In the collaboration setting, we use Listing \ref{listing:cf_code_collab} as the generator and Listing \ref{listing:cf_code_critic} as the critic.

Debugging is incorporated via a testing Flow that adds formatting to the output of a code executor. The formatting templates are shown in Listing \ref{listing:cf_code_testing}. To respond to the debug output, we rely on an adjusted coding Flow \ref{listing:cf_code_debug}.
Adding collaboration in the debugging setting is done by introducing a critic that provides feedback grounded in the test results. This Flow is detailed in Listing \ref{listing:cf_code_collab}.

The scenarios explained above also support the addition of a planning Flow. An example of plan generation is shown in Listing \ref{listing:cf_plan}.

\begin{lstlisting}[caption=Prompts for Code Flow (Codeforces), label={listing:cf_code}, breaklines]
"prompt templates":
  "system_message": |-
    Your goal is to provide executable Python code that solves a competitive programming problem. The code should correctly handle all corner cases in order to pass the hidden test cases, which are used to evaluate the correctness of the solution.

    The user will specify the problem by providing you with:
      - the problem statement
      - input description
      - output description
      - example test cases
      - (optional) explanation of the test cases

    The user will provide you with a task and an output format that you will strictly follow.
  "query_message": |-
    # Problem statement
    {{problem_description}}

    # Input description
    {{input_description}}

    # Output description
    {{output_description}}

    {{io_examples_and_explanation}}


    The input should be read from the standard input and the output should be passed to the standard output.
    Return Python code that solves the problem. Reply in the following format:
    ```python
    {{code_placeholder}}
    ```
  "human_message": |-
    {{query}}

\end{lstlisting}

\begin{lstlisting}[caption=Prompts for Fixed-Reply Flow, label={listing:fixed_reply_reflect}, breaklines]
"prompt templates":
  "fixed_reply": |-
    Consider the problem statement and the last proposed solution. Are you sure that the solution is provided in the requested format, and crucially, solves the problem?
    If that is not the case, provide the corrected version of the code in the following format:
    ```python
    {{python_code}}
    ```
    otherwise, reply:
    "Final answer."

\end{lstlisting}

\begin{lstlisting}[caption=Prompts for Code-Collab Flow (Codeforces), label={listing:cf_code_collab}, breaklines]
"prompt templates":
  "system_message": |-
    Your goal is to provide executable Python code that solves a competitive programming problem. The code should correctly handle all corner cases in order to pass the hidden test cases, which are used to evaluate the correctness of the solution.

    The user will specify the problem by providing you with:
      - the problem statement
      - input description
      - output description
      - example test cases
      - (optional) explanation of the test cases

    The user will provide you with a task and an output format that you will strictly follow.
  "query_message": |-
    # Problem statement
    {{problem_description}}

    # Input description
    {{input_description}}

    # Output description
    {{output_description}}

    {{io_examples_and_explanation}}


    The input should be read from the standard input and the output should be passed to the standard output.
    Return Python code that solves the problem. Reply in the following format:
    ```python
    {{code_placeholder}}
    ```
  "human_message": |-
    # Feedback on the last proposed solution
    {{code_feedback}}


    Consider the original problem statement, the last proposed solution and the provided feedback. Does the solution need to be updated? If so, provide the corrected version of the code in the following format:
    ```python
    {{code_placeholder}}
    ```
    otherwise, reply:
    "Final answer."

\end{lstlisting}

\begin{lstlisting}[caption=Prompts for Code-Collab-Critic Flow (Codeforces), label={listing:cf_code_critic}, breaklines]
"prompt templates":
  "system_message": |-
    Your goal is to identify potential issues with a competitive programming solution attempt.

    The user will specify the problem by providing you with:
      - the problem statement
      - input description
      - output description
      - example test cases
      - (optional) explanation of the test cases
      - a Python solution attempt

    Crucially, your goal is to correctly identify potential issues with the solution attempt, and not to provide the code implementation yourself.
    The user will provide you with a task and an output format that you will strictly follow.
  "query_message": |-
    # Problem statement
    {{problem_description}}

    # Input description
    {{input_description}}

    # Output description
    {{output_description}}

    {{io_examples_and_explanation}}

    # Python solution attempt:
    ```python
    {{code}}
    ```


    Consider the problem statement and the solution attempt. Are there any issues with the proposed solution or it is correct? Explain your reasoning very concisely, and do not provide code.
  "human_message": |-
    {{query}}

\end{lstlisting}

\begin{lstlisting}[caption=Prompts for Code-Debug Flow (Codeforces), label={listing:cf_code_debug}, breaklines]
"prompt templates":
  "system_message": |-
    Your goal is to provide executable Python code that solves a competitive programming problem. The code should correctly handle all corner cases in order to pass the hidden test cases, which are used to evaluate the correctness of the solution.

    The user will specify the problem by providing you with:
      - the problem statement
      - input description
      - output description
      - example test cases
      - (optional) explanation of the test cases

    The user will provide you with a task and an output format that you will strictly follow.
  "query_message": |-
    # Problem statement
    {{problem_description}}

    # Input description
    {{input_description}}

    # Output description
    {{output_description}}

    {{io_examples_and_explanation}}


    The input should be read from the standard input and the output should be passed to the standard output.
    Return Python code that solves the problem. Reply in the following format:
    ```python
    {{code_placeholder}}
    ```
  "human_message": |-
    {{testing_results_summary}}


    Consider the problem statement, the last proposed solution, and its issue. Provide a corrected version of the code that solves the original problem and resolves the issue, without any explanation, in the following format:
    ```python
    {{code_placeholder}}
    ```

\end{lstlisting}

\begin{lstlisting}[caption=Formatting templates for Code-Testing Flow (Codeforces), label={listing:cf_code_testing}, breaklines]
"formatting templates":
  "no error template": |-
    ${.issue_title}
    All of the executed tests passed.
  "all tests header": |-
    ${.issue_title}
    The Python code does not solve the problem in the problem description due to logical errors. It fails on the following tests.
  "compilation error template": |-
    ${.issue_title}
    The execution resulted in a compilation error.
    ## Compilation error message:
    {{error_message}}
  "timeout error template": |-
    ${.issue_title}
    The execution timed out, the solution is not efficient enough.
  "runtime error template": |-
    ${.issue_title}
    The execution resulted in a runtime error on the following test.
    ## [Failed test] Input
    ```
    {{test_input}}
    ```
    ## [Failed test] Runtime error message
    {{error_message}}
  "single test error": |-
    ${.issue_title}
    The Python code does not solve the problem in the problem description due to logical errors. It fails the following test:
    ## [Failed test] Input
    ```
    {{test_input}}
    ```
    ## [Failed test] Expected output
    ```
    {{expected_output}}
    ```
    ## [Failed test] Generated output
    ```
    {{generated_output}}
    ```
  "test error": |-
    ## [Failed test {{idx}}]
    ### [Failed test {{idx}}] Input
    ```
    {{test_input}}
    ```
    ### [Failed test {{idx}}] Expected output
    ```
    {{expected_output}}
    ```
    ### [Failed test {{idx}}] Generated output
    ```
    {{generated_output}}
    ```

\end{lstlisting}

\begin{lstlisting}[caption=Prompts for Code-Debug-Collab Flow (Codeforces), label={listing:cf_code_debug_collab}, breaklines]
"prompt templates":
  "system_message": |-
    Your goal is to identify the issues with an incorrect competitive programming solution attempt.

    The user will specify the problem by providing you with:
      - the problem statement
      - input description
      - output description
      - example test cases
      - (optional) explanation of the test cases
      - an incorrect Python solution attempt and a description of its issue

    Crucially, your goal is to consider all aspects of the problem and pinpoint the issues with the solution attempt, and not to provide the code implementation yourself.
    Some aspects to consider: Is the input correctly parsed? Is the output correctly formatted? Are the corner cases correctly handled? Is there a logical mistake with the algorithm itself?
    Use the code execution results provided in the issue description to guide your reasoning/debugging.
  "query_message": |-
    # Problem statement
    {{problem_description}}

    # Input description
    {{input_description}}

    # Output description
    {{output_description}}

    {{io_examples_and_explanation}}

    # Solution attempt to be fixed
    ```python
    {{code}}
    ```

    {{testing_results_summary}}


    Consider the problem statement, the solution attempt and the issue. Why is the solution attempt incorrect? How should it be fixed? Explain your reasoning very concisely, and do not provide code.
  "human_message": |-
    {{query}}

\end{lstlisting}

\begin{lstlisting}[caption=Prompts for Plan Flow (Codeforces), label={listing:cf_plan}, breaklines]
"prompt templates":
  "system_message": |-
    Your goal is to provide a high-level conceptual solution that, if implemented, will solve a given competitive programming problem.

    The user will specify the problem by providing you with:
      - the problem statement
      - input description
      - output description
      - example test cases
      - (optional) explanation of the test cases

    The proposed algorithm should be computationally efficient, logically correct and handle all corner cases.

    The user will provide you with a task and an output format that you will strictly follow.
  "query_message": |-
    # Problem statement
    {{problem_description}}

    # Input description
    {{input_description}}

    # Output description
    {{output_description}}

    {{io_examples_and_explanation}}


    Return a high-level conceptual solution that would solve the problem. Be very concise, and do not provide code.
    Reply in the following format:
    # Conceptual solution
    {{plan_placeholder}}
  "human_message": |-
    {{query}}

\end{lstlisting}

\FloatBarrier
\subsection{The CC-Flows-competition: a new form of competitive coding}
\label{app:cc-competition}

Solving competitive coding challenges is an eminently hard problem.
The solve rate of only 27\% by directly attempting the problem and 47\% by the best-performing code Flow, paired with a reliable automatic evaluation metric, make competitive programming an ideal benchmark for AI systems. Motivated by this, we propose a competition where instead of people, proposed Flows solve competitive programming problems.


The competition will leverage the comprehensive dataset of publicly available Codeforces problems and the open-source infrastructure for inference and testing used in the experiments, available at 
\url{https://github.com/epfl-dlab/cc_flows}. 
The competition will only include problems published after the knowledge-cutoff date of GPT-4. Furthermore, not to overload the Codeforces online evaluation infrastructure, we further filter this dataset to problems for which public and private tests are available, and the output format is compatible with our local code testing infrastructure. Codeforces ranks the difficulty of each problem from 800 to 2100. At the time of publishing, we have the following number of problems per difficulty (total of 416):
\begin{itemize}
    \item difficulty 800: 149
    \item difficulty 900 to 1500 (inclusive): 185
    \item difficulty 1600 to 2100 (inclusive): 82
\end{itemize}

We will curate a leaderboard of best-performing Flows that will be publicly available on FlowVerse and provide the predictions that reproduce the reported scores using the provided infrastructure. 

The data will be released and should be used in accordance with Codeforces' Terms and Conditions. Concretely, Codeforces prohibits the material from being sold, sublicensed, or commercialized. For more details, take a look at the project's GitHub page.



\end{document}